\documentclass[letterpaper, 10 pt, conference]{ieeeconf}  

\IEEEoverridecommandlockouts                              
\IEEEoverridecommandlockouts                              

\usepackage{graphics} 
\usepackage{epsfig} 
\usepackage{mathptmx} 
\usepackage{times} 
\usepackage{amsmath} 
\usepackage{amssymb}  
\usepackage{svg}
\usepackage{cite}
\usepackage{amsmath,amssymb,amsfonts}
\usepackage{algorithmic}
\usepackage{graphicx}
\usepackage{textcomp}
\usepackage{multirow}
\usepackage{booktabs}
\usepackage{xcolor}
\usepackage{bbding}
\usepackage{pifont}
\usepackage{bm}
\newcommand{\etal}{\textit{et al}.}
\usepackage{amsmath}
\usepackage{multirow}               
\usepackage{multicol}               
\usepackage{multirow}               
\usepackage{float}                  
\usepackage{makecell}               
\usepackage{booktabs}               
\usepackage{amsmath}
\usepackage{amssymb}
\usepackage{array}
\usepackage{utfsym}
\usepackage{colortbl}  
\usepackage{xcolor}
\usepackage{pifont}
\usepackage{hyperref}
\definecolor{turquoise}{cmyk}{0.65,0,0.1,0.3}
\definecolor{purple}{rgb}{0.65,0,0.65}
\definecolor{dark_green}{rgb}{0, 0.5, 0}
\definecolor{orange}{rgb}{0.8, 0.6, 0.2}
\definecolor{red}{rgb}{0.8, 0.2, 0.2}
\definecolor{darkred}{rgb}{0.6, 0.1, 0.05}
\definecolor{blueish}{rgb}{0.0, 0.3, .6}
\definecolor{light_gray}{rgb}{0.7, 0.7, .7}
\definecolor{pink}{rgb}{1, 0, 1}
\definecolor{greyblue}{rgb}{0.25, 0.25, 1}
\definecolor{igray}{gray}{.9}
\definecolor{pink}{rgb}{1, 0, 1}
\definecolor{pinkishred}{rgb}{0.9647058823529412, 0.6, 0.8196078431372549}
\definecolor{ForestGreen}{RGB}{21,155,82}

\def\etal{{\em et al.}}

\usepackage[cal=cm]{mathalfa}

\title{\LARGE \bf
Domain Adaptive Lung Nodule Detection in X-ray Image
}

\author{Haifeng Zhao, Lixiang Jiang, Leilei Ma, Dengdi Sun, and Yanping Fu
\thanks{All authors are with the Anhui Provincial Key Laboratory of Multimodal Cognitive Computation, School of Computer Science and Technology, Anhui University, Hefei, China.
Corresponding authors: Yanping Fu ({\tt\small ypfu@ahu.edu.cn}) and Leilei Ma ({\tt\small xiaomylei@163.com})
}}

\def\BibTeX{{\rm B\kern-.05em{\sc i\kern-.025em b}\kern-.08em
    T\kern-.1667em\lower.7ex\hbox{E}\kern-.125emX}}

\begin{document}

\maketitle
\thispagestyle{empty}
\pagestyle{empty}

\begin{abstract}
Medical images from different healthcare centers exhibit varied data distributions, posing significant challenges for adapting lung nodule detection due to the domain shift between training and application phases. Traditional unsupervised domain adaptive detection methods often struggle with this shift, leading to suboptimal outcomes. To overcome these challenges, we introduce a novel domain adaptive approach for lung nodule detection that leverages mean teacher self-training and contrastive learning. First, we propose a hierarchical contrastive learning strategy to refine nodule representations and enhance the distinction between nodules and background. Second, we introduce a nodule-level domain-invariant feature learning (NDL) module to capture domain-invariant features through adversarial learning across different domains. Additionally, we propose a new annotated dataset of X-ray images to aid in advancing lung nodule detection research. Extensive experiments conducted on multiple X-ray datasets demonstrate the efficacy of our approach in mitigating domain shift impacts.
\end{abstract}

\section{INTRODUCTION}
\begin{figure}[htp]
    \centering
    \includegraphics[width=0.4\textwidth]{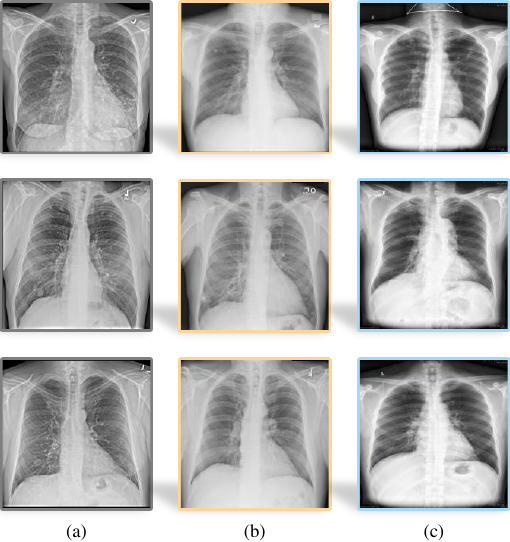}
    \caption{Samples from multiple X-ray datasets: (a) NODE21~\cite{10479589}, (b) CXR~\cite{cxr-dcjlk_dataset}, and (c) B-Nodule dataset, which is collected and annotated by ourselves. There exist certain domain differences among chest X-ray images from different datasets regarding illumination, color contrast/saturation, resolution, and nodule quantity.}
    \label{figure1}
\end{figure}

Lung nodules play a crucial role in indicating lung cancer, underscoring the paramount importance of early detection to improve patient survival rates~\cite{2schabath2019cancer}. Chest X-rays, being cost-effective and posing minimal radiation risk, are widely utilized by radiologists. However, the increasing workload poses challenges for physicians to effectively identify lesions in chest radiographs. Consequently, there is a pressing need for an automated lung nodule detection tool to assist doctors in swift and accurate diagnoses. In recent years, numerous researchers leverage object detection models for lung nodule detection~\cite{1tsai2022multi,3ji2023lung,4chen2020pulmonary,5shen2023image,li2020multi,sim2020deep}. Existing lung nodule detectors predominantly focus on data from the same medical center. Typically, training and test samples in these models originate from the same domain, which shares similar distributions. However, in many scenarios, training and application data may come from diverse centers, leading to a challenge known as the domain shift problem, as shown in Fig.~\ref{figure1}. This issue arises when a model trained on source domain data exhibits suboptimal performance when applied to target domain data. 

In recent years, domain adaptive object detection algorithms have garnered significant attention. Some approaches address the challenge of domain adaptation for object detection by mean teacher self-training~\cite{6cai2019exploring,7deng2021unbiased,9chen2022learning,10cao2023contrastive}. However, methods relying on mean teacher self-training often encounter the challenge of generating low-quality pseudo-labels. Since the knowledge of the teacher model is derived from training on the source domain data, the pseudo-labels generated by the teacher always contain noise. These noisy pseudo-labels, when used as supervision signals for the student model on target domain images, can adversely affect performance. The Probabilistic Teacher (PT)~\cite{9chen2022learning} mitigates the impact of noisy labels by capturing the uncertainty of unlabeled target data. The Contrastive Mean Teacher (CMT)~\cite{10cao2023contrastive} brings the features of the same class objects closer between the teacher model and the student model, yet it neglects the intrinsic data relationships within the student model. Other methods~\cite{12li2022cross,hsu2020every, li2022scan} leverage domain classifiers and adversarial networks to obtain domain-invariant features. The Adaptive Teacher (AT)~\cite{12li2022cross} incorporates domain classifier and adversarial network to alleviate shift. However, they often focus solely on image-level features, overlooking finer granularity in feature-level domain alignment. Approaches such as Every Pixel Matters (EPM)~\cite{hsu2020every} and Semantic Conditioned Adaptation (SCAN)~\cite{li2022scan} primarily concentrate on the variations between source and target domain data but often disregard additional informative elements, such as the relationship between the foreground and background, which could serve as valuable supervision signal for the model.

To address these deficiencies, we propose a novel lung nodule detection method that enhances mean teacher self-training with hierarchical contrastive learning and a nodule-level domain-invariant feature learning (NDL) module. The proposed approach tackles the challenge of low-quality pseudo-labels by employing a distinct hierarchical contrastive learning strategy. Unlike previous methods that primarily focus on data variations between source and target domains, our strategy also captures additional valuable information, such as the correlation between nodules and background, throughout the training process. Our hierarchical contrastive learning strategy unfolds in two dimensions: region-level and pixel-level. The region-level contrastive learning emphasizes the size and shape characteristics of objects in both nodule and background areas, whereas pixel-level contrastive learning targets the texture and grayscale features within these regions. Additionally, we integrate a nodule-level domain-invariant feature learning (NDL) module, which features an adversarial gradient reversal layer (GRL)~\cite{13ganin2015unsupervised} and a domain classifier, aimed at refining the model’s performance by capturing finer details and addressing domain adaptive challenges at the object level. To validate the effectiveness and generalization of the model, we constructed a new lung nodule detection dataset named B-Nodule. We collected the dataset from collaborative hospitals and annotated it by experienced radiologists. We validate our method through domain adaptive experiments for lung nodule detection across two public datasets and our dataset B-Nodule. The experimental results demonstrate that our approach effectively mitigates the domain shift issue and surpasses existing methods in performance.

In summary, our contributions consist of the following:
\begin{itemize}
\item We propose a novel lung nodule detection method based on the mean teacher approach and contrastive learning, which can effectively alleviate domain shifts in cross-domain lung nodule detection.
\item We introduce a hierarchical contrastive learning strategy, which is designed to enhance the distinctiveness of nodule representations at the region-level contrastive learning and pixel-level contrastive learning.
\item We propose a nodule-level domain-invariant feature learning (NDL) module, which incorporates an adversarial gradient reversal layer and a domain classifier to align nodule-level features across different domains.
\item We present a new annotated dataset of chest X-ray images for lung nodule detection called B-Nodule, which can effectively expand the datasets and promote further lung nodule detection research.
\end{itemize}

\section{RELATED WORK}
\subsection{Lung Nodule Detection in Chest X-ray image }

Lung nodule detection plays a crucial role in early lung cancer prevention and diagnosis. The utilization of chest X-ray images for examining lung nodules is a prevalent practice among radiologists, owing to its cost-effectiveness and widespread availability. With the advancement of deep learning technology, researchers have developed numerous automatic lung nodule detection algorithms. These algorithms serve as valuable tools to aid doctors in diagnosis, improving the efficiency and accuracy of lung nodule detection processes. Tsai~\etal~\cite{1tsai2022multi} propose a multi-task lung nodule detection algorithm using a Dual Head Network (DHN), which forecasts global-level labels for nodule presence and local-level labels for precise location. Additionally, Shen~\etal~\cite{5shen2023image} introduce an innovative lung nodule synthesis framework, leveraging data augmentation to address challenges posed by limited datasets. Chen~\etal~\cite{4chen2020pulmonary} apply a multi-segment active shape model for accurate lung parenchyma segmentation and utilize grayscale morphological enhancement to improve nodule structure visibility. While performing well in closed domains, these lung nodule detectors face challenges when it comes to generalizing to unseen domains.

\subsection{Unsupervised Domain Adaptation for Object Detection}
The unsupervised domain adaptation is originally researched in image classification~\cite{ganin2016domain,saito2018maximum,ghifary2016deep}, focusing on scenarios involving two domains: the source domain and the target domain, each with distinct data distributions. Models trained on the source domain often exhibit poor performance when applied in the target domain. Various approaches have been explored to tackle this issue.

Some methods emphasize domain-invariant feature learning, incorporating discriminators and adversarial training of feature encoders to achieve domain-invariant features~\cite{chen2018domain,saito2019strong,sindagi2020prior,hsu2020every,li2022scan,chen2021scale}. 
Others opt for image-to-image translation techniques, translating images from the target domain to match the style of the source domain, thereby addressing domain adaptive challenges~\cite{zhang2019cycle,chen2020harmonizing,hsu2020progressive,shen2021cdtd}. Moreover, some approaches integrate randomized domain features during training to mitigate domain shift problem~\cite{kim2019diversify,1rodriguez2019domain,tarvainen2017mean}. Additionally, The mean teacher, originally designed for semi-supervised problems, has been adapted to address unsupervised domain adaptive issues by~\cite{6cai2019exploring}. Following this exploration, such as Probabilistic Teacher (PT)~\cite{9chen2022learning}, Adaptive Teacher (AT)~\cite{12li2022cross}, Contrastive Mean Teacher (CMT)~\cite{10cao2023contrastive}, and Harmonious Teacher (HT)~\cite{11deng2023harmonious} have been introduced to enhance classification and localization accuracy in domain adaptive object detection. These models employ strategies like uncertainty-guided consistency training, adversarial enhancement, contrastive learning, and consistency regularization to improve overall performance. However, all these approaches are susceptible to a shared issue found in mean teacher methods, involving the generation of low-quality pseudo-labels in the target domain.

\subsection{Contrastive Learning}
Contrastive learning is a training strategy that involves learning feature representations by comparing similar and dissimilar data pairs, aiming to bring similar samples closer in the feature space while pushing dissimilar samples farther apart. Recently, various computer vision tasks, including image classification~\cite{ma2023semantic,ma2024text}, object detection~\cite{henaff2021efficient}, image segmentation~\cite{van2021unsupervised}, and more have achieve batter performance by incorporating contrastive learning strategies. Morever, The Contrastive Mean Teacher (CMT)~\cite{10cao2023contrastive} alleviates the domain shift in object detection through contrastive learning across the teacher model and student model. Inspired by these approaches, our hierarchical contrastive learning strategy leverages region-level and pixel-level contrastive learning to obtain more discriminative nodule and background representations, thereby achieving more accurate cross-domain lung nodule detection.

\section{METHOD}
\begin{figure*}[htp]
    \centering
    \includegraphics[width=1.0\textwidth]{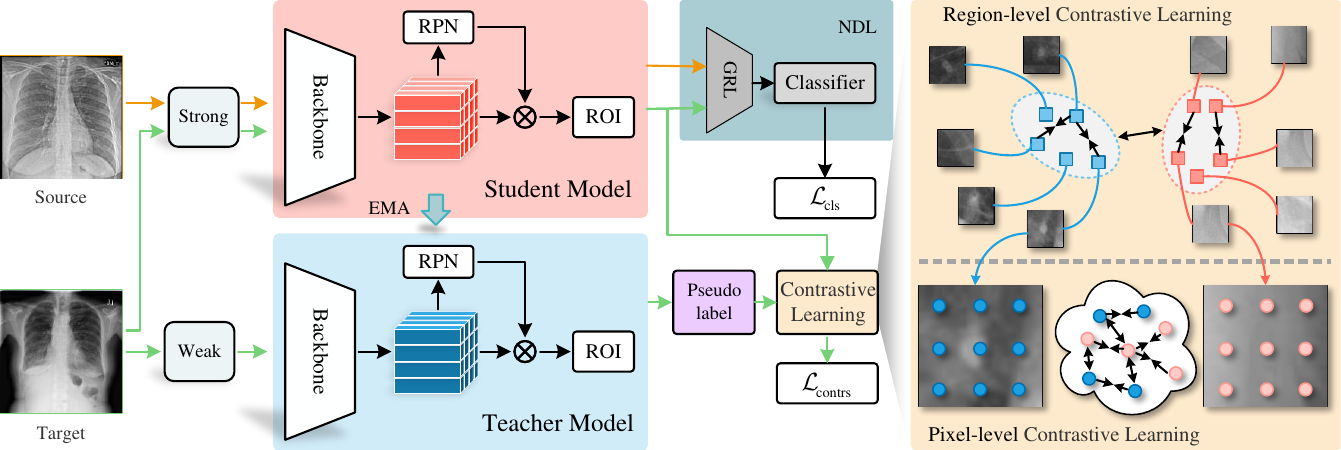}
    \caption{\textbf{Overview of our model}. \textbf{Left:} Teacher-student mutual learning and nodule-level domain-invariant feature learning (NDL) module. The teacher provides pseudo-labels to supervise the student, and the student refines the teacher by Exponential Moving Average (EMA). \textbf{Right:} Hierarchical contrastive learning strategy.} 
    \label{figure2}
\end{figure*}

Our goal is to address the challenge of domain adaptation for lung nodule detection. In the following sections, we provide a detailed description of our proposed model architecture, as shown in Fig.~\ref{figure2}. We first integrate hierarchical contrastive learning into the mean teacher model. We design a novel hierarchical contrastive learning strategy, which involves region-level and pixel-level contrastive learning, utilizing the differences between foreground and background as supervisory signals for target data. Then, we introduce a nodule-level domain-invariant feature learning (NDL) module to align the two domains, further enhancing the adaptation of the model to the target domain. Finally, we introduce teacher-student mutual learning, where the weight of the student model is used to update the teacher model, improving the accuracy in detecting pseudo-nodules.

\subsection {Hierarchical Contrastive Learning}\label{AA}
To improve the quality of pseudo nodules generated during training of teacher model on target domain, inspired by~\cite{xie2022c2am,joseph2021towards}, we propose hierarchical contrastive learning into the proposed model. The hierarchical contrastive learning strategy encompasses two levels: region-level contrastive learning and pixel-level contrastive learning, as shown in Fig.~\ref{figure2}.

\textbf{Region-level Contrastive Learning}. We classify the region proposals generated from the student model into two categories: positive (nodule) and negative (background) regions, employing intersection over union (IOU) pseudo-labels as the criteria. Utilizing contrastive learning, we bring together foreground and foreground, background and background, while creating a clear distinction between foreground and background elements. This process ensures a more distinct representation of nodules, significantly improving the capability of the detector to distinguish between foreground and background.

Assuming that the region proposal network generates n region proposals, the features of these regions are denoted as $S_i$, where $i \in \{ 1,2,\ldots,n \}$. These features of these regions are classified into two categories, positive $S_{1:j}^+$, negative $S_{1:k}^-$ with $j+k=n$ by pseudo-labels. The region contrastive loss is defined as follows:
\begin{equation}
\mathcal{L}_{\text{region}}=D(S_{1:j}^+,S_{1:k}^-)+\overline{D}(S_{1:j}^+)+\overline{D}(S_{1:k}^-)~\text{,}
\end{equation}
where $D\left(\cdot, \cdot\right)$ is the function that separates the positive region from the negative region, while $\overline{D}\left(\cdot\right)$ is the function that brings samples of the same class closer to another.

\textbf{Pixel-level Contrastive Learning}. The region-level contrastive learning strategy concentrates on aspects like nodule size and morphology but misses finer characteristics. Therefore, we introduce pixel-level contrastive learning, which highlights details such as color and texture, facilitating a more nuanced distinction between nodules and background in the representation.

To ensure that the pixels come from salient areas, we sample the top p positive pixel features $s_{1:p}^+$ and the top q negative pixel features $s_{1:q}^-$ based on the sum of the channel values from the region-level features $S_{1:j}^+$ and $S_{1:k}^-$. Subsequently, using pixel-level contrastive loss, we bring positive pixels closer to each other positive pixels and negative pixels closer to each other negative pixels, while pushing positive pixels away from negative pixels. The pixel-level contrastive loss is defined as follows:
\begin{equation}
	\mathcal{L}_{\text{pixel}}=D(s_{1:p}^+ , s_{1:q}^-)+\overline{D}(s_{1:p}^+)+\overline{D}(s_{1:q}^-)~\text{.}
\end{equation}

Finally, the hierarchical contrastive loss is composed of both pixel-level loss and region-level loss working together.
\begin{equation}
	\mathcal{L}_{\text{contrs}}=\lambda \mathcal{L}_{\text{region}}+(1-\lambda) \mathcal{L}_{\text{pixel}}~,
\end{equation}
where $\lambda$ is the hyperparameter to balance two losses.

The aim of the proposed method is to make nodules more alike to each other in representation while making them more distinct from the background. This strategy is designed to boost the overall similarity among nodule representations and sharpen their contrast with background features. Thus, the function $D\left(\cdot, \cdot\right)$ can be defined as 
\begin{equation}
D(A,B)=-\frac{1}{\left | A \right | \left | B \right |} \sum_{a \in A}^{} \sum_{b\in B}^{}\log_{}{(1-\text{sim}(a,b))}~,
\end{equation}
where $A$ and $B$ represent the sets of region-level or pixel-level features, the $\text{sim}\left(\cdot, \cdot\right)$ denotes the cosine similarity between the feature $a$ and feature $b$.

However, lung nodules can display entirely diverse feature representations due to variations in structural morphology across different types of lung nodules. Moreover, there are distinctions in feature representations within backgrounds. Therefore, indiscriminately converging samples of the same category in feature representation is detrimental to model training. To tackle this challenge, we employ feature similarity ranking to adjust the weighting of similarity losses. The loss function $\overline{D}\left(\cdot\right)$ for bringing samples of the same class closer is defined as  
\begin{equation}
\overline{D}(A)\! =\!-\frac{1}{\! \left | A \right |( \left | A \right |\!-\!1)} \sum_{a_1\in A}^{}\sum_{a_2\in A}^{} \mu_{a_1,a_2} \text{sim}(a_1,a_2)_{(a_1\ne a_2)}~,
\end{equation}
where the weight $\mu_{a_1,a_2}$ of each sample pair is obtained by ranking the similarity of the sample pairs.
\begin{equation}
\mu_{a_1,a_2}=exp(-{\mu} * \text{rank}(\text{sim}(a_1,a_2)))~,
\end{equation}
where $\mu$ represents a smoothing hyperparameter for the modulated exponential function. The proposed method assigns a higher weight to the loss from pairs of samples that are more similar and a smaller weight to the loss from pairs of samples that are less similar. This mechanism ensures improved contrastive learning by emphasizing more meaningful sample pairs while downplaying less relevant ones.

\subsection{NDL Module}\label{BB}
The teacher and student models are initially trained using data from the source domain. However, because the source and target domains have inherent differences, there is a domain shift in the feature maps between them. During the mutual learning process, the teacher model generates pseudo-labels based on knowledge derived from the source domain, which may be of lower quality. Utilizing these lower-quality pseudo-labels to supervise the training of the student model can be problematic. To mitigate this issue, we focus on aligning the nodule features between the two domains.

Lung nodules in X-ray images from distinct domains may vary in size, structure, texture, and other aspects. In the lung nodule detection model, we utilize features at the object level obtained through Region of Interest (ROI) pooling for classification and localization. Therefore, unlike previous approaches~\cite{12li2022cross,saito2019strong} that typically concentrate on aligning source and target domains at the image level, our approach emphasizes the alignment of more detailed features, specifically, the alignment of nodule-level features. Therefore, we propose a nodule-level domain-invariant feature learning (NDL) module to align the two domains.

After obtaining nodule-level features from the source and target domains following the ROI process, we feed them into a domain classifier. This classifier provides the probability of the feature belonging to the source domain $P$ and the target domain $(1-P)$. Each nodule-level feature is associated with a domain label $T$. For source domain images, $T=0$; and for target domain images, $T=1$. Using these domain labels and probability values, we calculate classifier loss $L_{cls}$ as
\begin{equation}
\mathcal{L}_{\text{cls}}=-T\log_{}{P} -(1-T)\log_{}{(1-P)}.
\end{equation}
To align features from both source and target domains and to hinder the domain classifier from making accurate predictions, we introduce an adversarial gradient reversal layer (GRL) before the domain classifier. The domain classifier is optimized based on $\mathcal{L}_{\text{cls}}$ during mutual learning. Throughout forward propagation, the adversarial gradient reversal layer leaves the input unchanged, while during reverse propagation, it inversely multiplies the gradient by a negative scalar. This process encourages the encoder to generate nodule-level features that are challenging to distinguish, thereby addressing the domain adaptive challenge in lung nodule detection.

\subsection{Teacher-Student Mutual Learning}\label{CC}
In the domain adaptive training of mean teacher model, our method starts by training an adapted model using data from the source domain and obtain the weight of the adapted model $\theta$. Subsequently, we replicate the model to establish a teacher-student pair. During training, pseudo-labels generated by the teacher play a crucial role in guiding the student model. The teacher undergoes incremental updates using the student weights and follows an exponential moving average (EMA) approach. In this way, it can be considered as multiple temporal ensemble student models. The weight of teacher model is updated for each time step as
\begin{equation}
{\theta_t} \leftarrow \beta {\theta _t}+(1-\beta ){\theta_s}~,
\end{equation}
where $\theta_t$ and $\theta_s$ denote the weights of teacher model and student model respectively.

\section{EXPERIMENTS}
\subsection{Experimental Settings}
\textbf{Datasets.} We conduct experiments on three datasets, including NODE21\footnote{\url{https://node21.grand-challenge.org/Data/}}~\cite{10479589}, CXR\footnote{\url{https://universe.roboflow.com/xray-chest-nodule/cxr-dcjlk}}~\cite{cxr-dcjlk_dataset}, and our dataset B-Nodule\footnote{\url{https://github.com/yfpeople/B-nodule-Dataset}}. We collect and annotate a lung nodule in an X-ray image dataset named B-Nodule, which comprises $1964$ chest X-ray images and $2314$ annotated lung nodules. This dataset is split into training and test sets to facilitate robust model evaluation. For training, the B-Nodule dataset is annotated by professional radiologists. 

\textbf{Implementation Details.}
We employ Faster R-CNN~\cite{ren2015faster} as the base detector. The pretrained VGG16~\cite{DBLP:journals/corr/SimonyanZ14a} is used as the backbone. All chest X-ray images are resized to $640\times640$. We set the batch size to $16$. We set the region of interest (ROI) threshold to $0.75$ and the balanced parameter $\lambda$ for hierarchical contrastive loss to $0.7$. The model is trained on the source labels for 10,000 iterations, including 2,000 iterations for pretraining and 8,000 iterations for self-learning. The learning rate for the entire training phase is set to 0.04, with no application of any learning rate decay. We utilized stochastic gradient descent to optimize the network, and the weight smoothing coefficient for the exponential moving average (EMA) of the teacher model is set to $0.9996$.

\textbf{Evaluation Metrics.} The primary metric used to evaluate detection performance is the mean average precision. The average precision for each threshold is determined by calculating the area under the precision-recall curve, which is interpolated at $101$ points. The mean average precision ($\text{AP}$) is computed as the average of the average precision over IOU thresholds ranging from $0.5$ to $0.95$, in increments of $0.05$. $\text{AP}_{50}$ refers to the average precision at an IOU threshold of $0.5$, $\text{AP}_{75}$ refers to the average precision at an IOU threshold of $0.75$. Additionally, $\text{AP}_{S}$ denotes the mean average precision for small nodules (area less than $32\times 32$), $\text{AP}_{M}$ for medium nodules (area between $32\times 32 $ and $96\times 96$), and $\text{AP}_{L}$ for large nodules (area greater than $96\times 96$).


\begin{table}[t]
    \centering{
    \renewcommand{\arraystretch}{1.25}
    \caption{Comparative results on the experimental scenario where the setting is at CXR to B-Nodule.}
    \resizebox{\linewidth}{!}{
    \begin{tabular}{c | c  c  c  c  c}
        \toprule
        \hline
        Method &  $\text{AP}$ & $\text{AP}_{50}$ & $\text{AP}_{75}$ &  $\text{AP}_{M}$ & $\text{AP}_{L}$ \\
        \hline
        EPM\cite{hsu2020every} & 18.074 & 50.874 & 4.689 & 12.951 & 19.531 \\
        SCAN\cite{li2022scan} &   19.960 & 49.380 & 9.438 &\textbf{22.389} & 20.561 \\
        PT\cite{9chen2022learning} &  20.879 & 53.791 & 8.828 &  17.944 & 21.940 \\
        AT\cite{12li2022cross} &  14.542 & 43.759 & 4.025 &  7.709 & 16.014 \\
        CMT\cite{10cao2023contrastive} & 20.591 & 52.679 & 9.195 &  12.698 & 22.000 \\
        \hline
        PT+Ours &  \textbf{21.928} & \textbf{56.337} & \textbf{9.730} & 14.370 & \textbf{23.659} \\
        \hline
        \bottomrule
    \end{tabular}
	}
	\label{tab1}
    }
\end{table}
\begin{table}[t]
    \centering{
    \renewcommand{\arraystretch}{1.25}
    \caption{Comparative results on the experimental scenario where the setting is at CXR to NODE21.}
    \resizebox{\linewidth}{!}{
    \begin{tabular}{c | c  c  c  c  c c}
        \toprule
        \hline
        Method &  $\text{AP}$ & $\text{AP}_{50}$ & $\text{AP}_{75}$ & $\text{AP}_{S}$ & $\text{AP}_{M}$ & $\text{AP}_{L}$ \\
        \hline
        EPM~\cite{hsu2020every} &  17.526 & 50.250 & 4.743 & 0.495 & 18.568 & 26.341 \\
        SCAN~\cite{li2022scan} &  16.790 & 50.439 & 5.510 & 3.267 & 18.002 & 16.507 \\
        PT~\cite{9chen2022learning}&  17.784 & 48.158 & 8.452 & 4.093 & 18.386 & 26.415 \\
        AT~\cite{12li2022cross} & 18.137 & 45.484 & 9.915 & 4.770 & 18.310 & 27.077 \\
        CMT~\cite{10cao2023contrastive} &   19.508 & 49.865 & 10.237 & 5.643 & 19.921 & \textbf{30.814} \\
        \hline
        PT+Ours &  \textbf{19.948} & \textbf{50.671} & \textbf{10.355} & \textbf{5.755} & \textbf{20.889} & 28.493 \\
        \hline
        \bottomrule
    \end{tabular}
	}
	\label{tab2}
    }
\end{table}

\subsection{Results and Comparisons}
In this section, we evaluate our proposed method against several established domain adaptive object detection methods in two scenarios: CXR to B-Nodule and CXR to NODE21. The methods we compare against include EPM~\cite{hsu2020every}, SCAN~\cite{li2022scan}, PT~\cite{9chen2022learning}, AT~\cite{12li2022cross}, and CMT~\cite{10cao2023contrastive}, all of which are prominent in the field of unsupervised domain adaptation for object detection.

\textbf{CXR to B-Nodule.}
By incorporating the Probabilistic Teacher (PT) approach into our framework, we achieve significant performance improvements, obtaining the best results, as shown in TABLE~\ref{tab1}. We conducted comparisons between our method and various other domain adaptive object detection approaches, particularly emphasizing the performance differences of each competing method in the CXR to B-Nodule scenario. It is evident that our method outperforms all others, including the nearest competitor, the Probabilistic Teacher (PT), by a substantial margin. Our model achieved an $\text{AP}$ of 21.928 and an $\text{AP}_{50}$ of 56.337. Since all images in the B-nodule dataset are high-resolution, the lung nodule annotations in this dataset include only large nodules and a few medium nodules. Contrastive learning and adversarial learning typically require adequate data samples to ensure the stability and effectiveness of the learning process. When there is a scarcity of annotations for medium nodules, the model struggles to capture sufficient intra-class variation and inter-class differences, resulting in suboptimal detection outcomes, which is reflected in a lower $\text{AP}_m$. However, for larger nodules with more annotations, the results in TABLE~\ref{tab1} highlight the superior performance of our model, further emphasizing the effectiveness of our approach.

\textbf{CXR to NODE21.}
Given that both the CXR and B-Nodule datasets include annotations for lung nodules across all images, we facilitated comparison in our study from CXR to NODE21 by exclusively using images from the NODE21 dataset that have annotated nodules. The results of setting CXR to NODE21 are presented in TABLE~\ref{tab2}. Notably, for small, medium, and large nodules, our method exhibits better performance compared to the original PT, consistently achieving higher $\text{AP}$. Furthermore, our model outperforms other methods across the $\text{AP}$, $\text{AP}_{50}$, and $\text{AP}_{75}$ metrics.

\begin{figure}[t]
    \centering
    \includegraphics[width=0.48\textwidth]{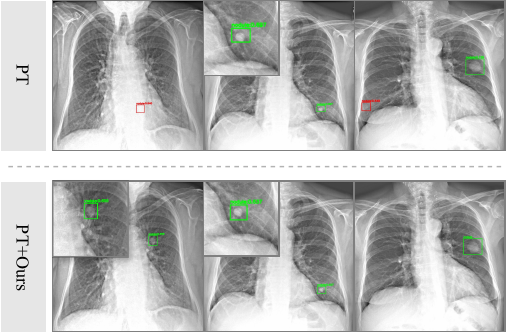}
    \caption{The comparison of the detection results of lung nodules between the approaches and our method in the scenario of CXR to NODE21. The small image serves as a magnified view of a specific region. The red boxes and green boxes denote the false positives (FP) and true positives (TP). }
    \label{figure3}
\end{figure}

\textbf{Qualitative Comparison.}
In a visual demonstration, we provide evidence of the superiority of our proposed method. In Fig.~\ref{figure3}, we present the performance of PT and our method in domain adaptive lung nodule detection from CXR to NODE21. As illustrated in Fig.~\ref{figure3}, our model effectively mitigates false positives. Furthermore, our method demonstrates superior performance in true positive detection, achieving both higher confidence scores and more accurate localization.
\begin{table}[t]
\renewcommand\arraystretch{1.15}
\caption{Ablation study of components of our method.}
\begin{center}
\resizebox{\linewidth}{!}{
\begin{tabular}{ c | c c  c | c c}
    \toprule
    \hline
    \multirow{2}{*}{Setting}&\multicolumn{2}{c}{\underline{Contrastive Learning}}& \multirow{2}{*}{$\mathcal{L}_{\text{cls}}$} &  \multirow{2}{*}{$\text{AP}$} & \multirow{2}{*}{$\text{AP}_{50}$}  \\
    & $\mathcal{L}_{\text{region}}$ &  $\mathcal{L}_{\text{pixel}}$ & & & \\
    \hline
    PT  &\textcolor{red}{\usym{2717}}         &\textcolor{red}{\usym{2717}}         &\textcolor{red}{\usym{2717}} & 20.879 & 53.791 \\
    (a) &\textcolor{ForestGreen}{\usym{2713}} & \textcolor{red}{\usym{2717}}        &\textcolor{red}{\usym{2717}} & 21.012 & 54.265\\
    (b) &\textcolor{ForestGreen}{\usym{2713}} &\textcolor{ForestGreen}{\usym{2713}} &\textcolor{red}{\usym{2717}} & 21.557 & 55.417  \\
    (c) &\textcolor{ForestGreen}{\usym{2713}} &\textcolor{ForestGreen}{\usym{2713}} &\textcolor{ForestGreen}{\usym{2713}} & \textbf{21.928} & \textbf{56.337}  \\
    \hline
    \bottomrule
\end{tabular}
}
\label{tab3}
\end{center}
\end{table}

\subsection{Ablation Study}
In this section, we conduct ablation studies on the critical components of the model. Specifically, we evaluate our model on CXR to B-Nodule, where the object detector is trained on CXR and tested on B-Nodule. In TABLE~\ref{tab3}, setting (a) and setting (b) represent the results of PT equipped with region-level contrastive loss and PT equipped with both region-level and pixel-level contrastive loss. Setting (c) shows the results of simultaneously employing contrastive learning and domain-invariant feature learning.

\textbf{Contrastive Loss.} 
The contrastive loss is subdivided into region-level contrastive loss and pixel-level contrastive loss. TABLE~\ref{tab3} demonstrates that upon integrating region-level contrastive loss, the $\text{AP}$ and $\text{AP}_{50}$ of the detector show improvements of 0.678 and 1.626. Additionally, the incorporation of pixel-level contrastive loss further enhancements of 0.545 $\text{AP}$ and 1.152 $\text{AP}_{50}$. This demonstrates the effectiveness of two level contrastive learning strategies. 

\textbf{Classifier Loss.}
We also validate the effectiveness of the NDL module. PT achieves the best results when both contrastive learning and NDL module are integrated, resulting in 0.371 $\text{AP}$ and 0.92 $\text{AP}_{50}$ improvement compared to using contrastive learning alone. This indicates that the NDL module contributes to improving the performance of the model.

\section{CONCLUSION}
In this paper, we propose a novel domain adaptive method for lung nodule detection in X-ray images. We tackle the challenge of domain shift through the integration of contrastive learning and domain-invariant feature learning. The contrastive learning approach, spanning from region to pixel scales, facilitates the model in acquiring meaningful representations of both nodules and backgrounds. Additionally, nodule-level domain-invariant feature learning aligns features between the source and target domains, thereby further mitigating the issue of low-quality pseudo-labels. The effectiveness of our approach is demonstrated through cross-domain experiments conducted on three datasets.

\section{ACKNOWLEDGEMENT}
This study is funded by the National Natural Science Foundation of China (62076005, U20A20398), the Natural Science Foundation of Anhui Province (2008085MF191, 2308085MF214), and the University Synergy Innovation Program of Anhui Province, China (GXXT-2021-002, GXXT-2022-029).
We thank the Hefei Artificial Intelligence Computing Center of Hefei Big Data Asset Operation Co., Ltd. for providing computational resources for this project.
\addtolength{\textheight}{-7cm}   
\bibliographystyle{IEEEtran}  
\bibliography{IEEEabrv,references}
\end{document}